\documentclass[runningheads]{llncs}
\usepackage{graphicx}
\usepackage{booktabs}
\newcommand{\keywords}[1]{\par\addvspace\baselineskip
\noindent\keywordname\enspace\ignorespaces#1}

\begin{document}

\title{Detection of Pitt-Hopkins Syndrome based on morphological facial features}
\titlerunning{Detection of Pitt-Hopkins Syndrome}

\author{Elena D'Amato \inst{1} \and Constantino Carlos Reyes-Aldasoro \inst{1,4} \and  Maria Felicia Faienza \inst{2} \and Marcella Zollino \inst{3}}

\authorrunning{D'Amato et al.}

\institute{School of Mathematics, Computer Science and Engineering\\
City, University of London, London, EC1V 0HB, UK \\
\and
Department of Biomedical Sciences and Human Oncology, Paediatric Section, \\
University of Bari  ``A. Moro", Bari, Italy\\
\and
IRCCS Fondazione Policlinico Universitario A. Gemelli, Rome, Italy
\and
Correspondence to: \email{reyes@city.ac.uk}
}

\maketitle

\begin{abstract}
This work describes an automatic methodology to discriminate between individuals with the genetic disorder  Pitt-Hopkins syndrome (PTHS), and healthy individuals.  As input data, the methodology accepts unconstrained frontal facial photographs, from which faces are located with Histograms of Oriented Gradients features descriptors. Pre-processing steps of the methodology consist of colour normalisation, scaling down, rotation, and cropping in order to produce a series of images of faces with consistent dimensions.  Sixty eight facial landmarks are automatically located on each face through a cascade of regression functions learnt via gradient boosting to estimate the shape from an initial approximation. The intensities of a sparse set of pixels indexed relative to this initial estimate are used to determine the landmarks. A set of carefully selected geometric features, for example, relative width of the mouth, or angle of the nose, are extracted from the landmarks. The features are used to investigate the statistical differences  between the two populations of PTHS and healthy controls.  The methodology was tested on 71 individuals with PTHS and 55 healthy controls. Two geometric features related to the nose and mouth showed statistical difference between the two populations. \\
\keywords{Pitt-Hopkins Syndrome, Morphological Face Analysis, Facial Landmarks}
\end{abstract}

\section{Introduction}
Pitt-Hopkins syndrome (PTHS) is a neurodevelopmental disorder caused  by haploinsufficiency of the TCF4 gene on chromosome 18, as a consequence of true deletion or intragenic mutations
\cite{Marangi_Zollino_2015,Goodspeed_Newsom_Morris_Powell_Evans_Golla_2018}. The individuals who are affected by PTHS  are normally characterised by an intellectual disability linked with developmental delay, unusual patterns of breathing, like breath-holding while awake or hyperventilation, epilepsy, and a distinctive facial gestalt. Some of the facial characteristics include sunken eyes, a squared forehead, larger than normal nose with a broad nasal bridge, a nasal tip that is arched with flaring nostrils, a noticeable curved shape of the upper lip (also known as Cupid's bow), and wide spaces between teeth \cite{Goodspeed_Newsom_Morris_Powell_Evans_Golla_2018,Amiel_Rio_de,Brockschmidt_Todt_Ryu_Hoischen_Landwehr_Birnbaum_Frenck_Radlwimmer_Lichter_Engels,Sweetser_Elsharkawi_Yonker_Steeves_Parkin_Thibert_1993}. 
Diagnosis of syndromes such as PTHS with genetic tests can be expensive, time-consuming and not available to some communities \cite{Kuru_Niranjan_Tunca_Osvank_Azim_2014}, and thus, an automatic methodology that can correlate genotype and phenotype and  pre-diagnose individuals with PTHS based on facial metrics  extracted from a single photograph is attractive.

The recognition of a genetic syndrome recognition based on a photograph is rather similar to the problem of classic facial recognition \cite{Samal_Iyengar_1992}. However,  a genetic syndrome detection will be limited by the amount of data available and the unbalance against  controls \cite{Schroff_Kalenichenko_Philbin_2015}, the difficulty in differentiating between subtle facial patterns, gender and ethnic differences \cite{Farnell_Galloway_Zhurov_Richmond_Perttiniemi_Katic_2017}, among other factors \cite{Gurovich_Hanani_Bar_Fleischer_Gelbman_Basel-Salmon_Krawitz_Kamphausen_Zenker_Bird}. Furthermore, the limited amount of data restricts the possibility of developing solutions based on deep learning approaches as these require very large amounts of training data. 

This paper describes a methodology that process unconstrained frontal facial photographs. The methodology extracts morphological measurements from geometric features through a series of image processing steps.
For anonymity purposes, the faces of the document have been obscured, and only the edges have been retained to illustrate the process. The actual processing was applied to the photographs.
These measurements showed statistical difference between individuals with PTHS and healthy controls. The methodology does not require user intervention nor the use of a large database. Although the methodology was tested with a relatively small database, the results are encouraging.

\section{Materials and Methods}
\subsection{Database}
A total of 126 photographs of Caucasian individuals aged between 2 and 17 years divided in two groups were analysed. The first group consisted of 71 cases of patients with a confirmed diagnosis of PTHS, which were selected from the Institute of Genomic Medicine, Gemelli Hospital Foundation, Rome, Italy. The second group consisted of 55 healthy controls acquired at Department of Biomedical Science and Human Oncology, University of Bari, Italy.
Written informed consent was signed by parents of the patients and healthy children. 
The photographs were acquired with a range of devices, mostly amateur cameras and smartphones, with variations of lighting, pose, distance from camera, background, orientation of device and resolution, the format for all of them was 8-bit RGB JPEG. All of the photographs were front or near-front facial images without  occlusions or partial view of the face.

\subsection{Methodology}
All processing was performed using Python 2.7 and the following libraries:  $sys, os, glob, Dlib$\cite{King_2009}, $NumPy$ \cite{Oliphant_2015}, $Scikit-image$\cite{vanderWalt_Schonberger}, $Scikit-learn$ \cite{Pedregosa_Varoquaux_Gramfort_Michel_Thirion_Grisel_Blondel_Prettenhofer_Weiss_Dubourg}. \\

The methodology consists of six stages. \\

\begin{figure}[!ht]
\centering
\begin{tabular}{cc}
{\includegraphics[width=59mm,height=64mm]{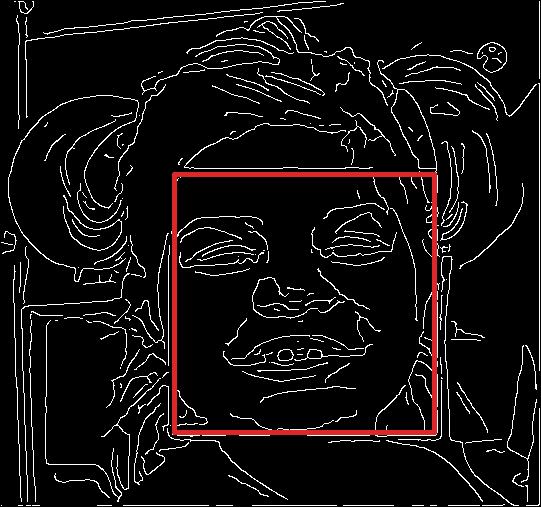}}&
{\includegraphics[width=59mm,height=64mm]{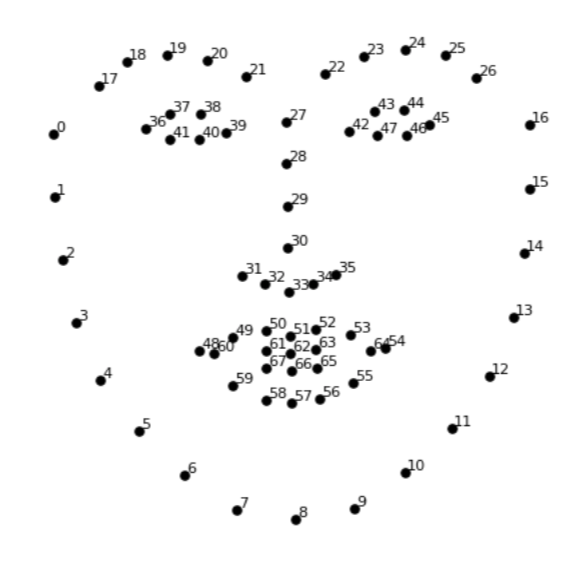}}\\
(a) & (b) \\
\end{tabular}
\caption{(a) One representative photograph of an individual with diagnosed PTHS. The face has been obscured for anonymity purposes. Canny edge detection was applied to the blue channel,  the regions without edges were set to black. Red bounding box denotes the region of interest that will be cropped. (b) Sixty eight facial landmarks to be localised on the photographs.}
\label{fig:figure1}
\end{figure}

1) {\bf Pre-processing}: Images were normalised using automatic histogram equalisation and median filtering in order to obtain uniform brightness and contrast levels. Faces were detected using the Viola-Jones object detection framework \cite{Viola_Jones_2004}. This algorithm detects faces by obtaining the sum of the values in image pixels within rectangular areas and then derive a large amount of features that are used to identify characteristics of a human face, e.g. the pixels of the eyes are darker than the pixels of the upper-cheeks. The location of the face was extended by 30 pixels on all directions and all pixels outside this region were discarded (Fig.~\ref{fig:figure1}(a)). This region with the face was the re-sized to $600\times600$ pixels. \\

2) {\bf Face detection}: A refined face detection process was then performed with  Histogram of Oriented Gaussians (HOG) \cite{Dalal_Triggs_2005}, which can provide better results than Viola-Jones \cite{Deniz_Bueno_Salido_DelaTorre_2011}. This was important for the location of landmarks. The HOG face detector assumes that  shape and texture of a face can be described by the distribution of gradients of intensity or edge directions. As first step of the HOG face detector, gradient vector (magnitude and angle) of every pixel of the picture is computed. The final HOG descriptor size is the vector of all components of the normalised blocks in the detection window. For this implementation, the size of the detection window was $64\times64$ pixels  
and $1400$ histograms or training vectors were generated.
The combined vectors are the input of a linear kernel support vector machine (SVM) \cite{Cortes_Vapnik_1995} classifier trained for face detection. The classifiers returned the bounding box of the face. \\ 

3) {\bf Face alignment and registration}: In the face alignment step, the face is rotated, scaled and translated taking into account the rigid and non-rigid face deformation. For this task, 5-point facial landmarks were used: 2 points for the corners of left eye, 2 points for the corners of the right eye and 1 point for the tip of the nose. The horizontal and vertical ($x, y$) coordinates of the five points were detected with a pre-trained model based on HOG descriptor and  SVM classifier. First of all, the centre of each eye (centroid) was calculated in accordance with the two corner points detected. Then, taking into account  the differences in $x$ and $y$ coordinates of the centroids of both eyes and arc-tangent function, the angle of rotation was calculated. Once the face is rotated, it is scaled back to the dimension of 600x600 pixels. \\

\begin{figure}[!ht]
\centering
\begin{tabular}{cc}
{\includegraphics[width=59mm,height=64mm]{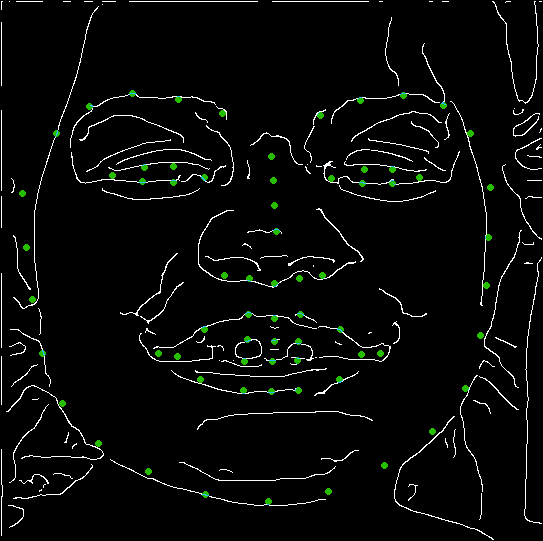}}&
{\includegraphics[width=59mm,height=64mm]{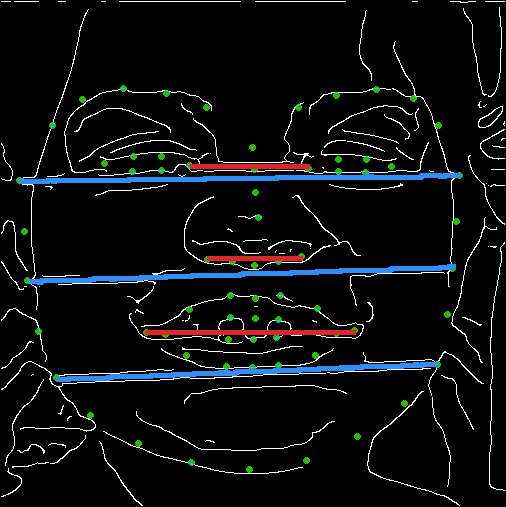}}\\
(a) & (b) \\
{\includegraphics[width=59mm,height=64mm]{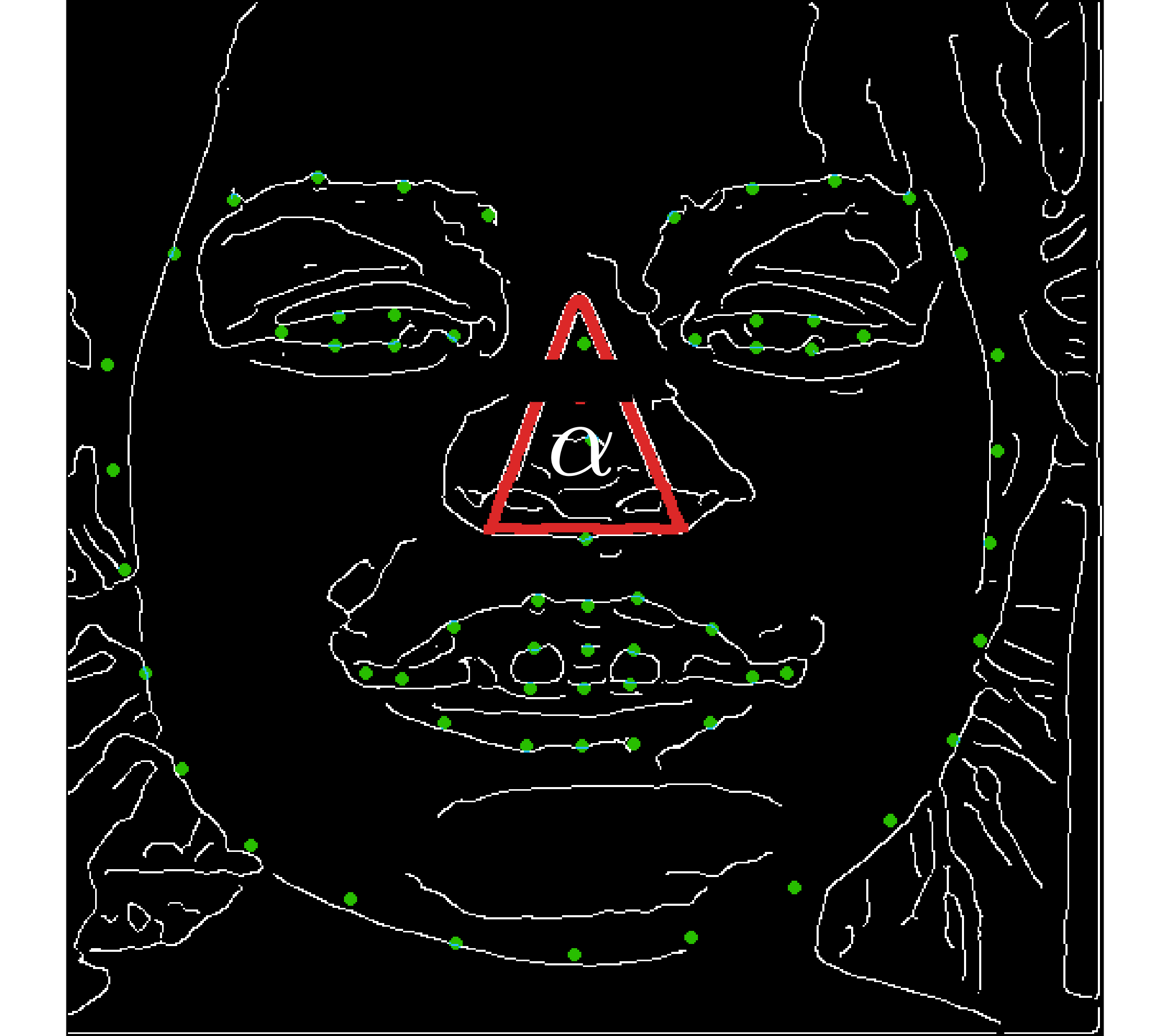}}&
{\includegraphics[width=59mm,height=64mm]{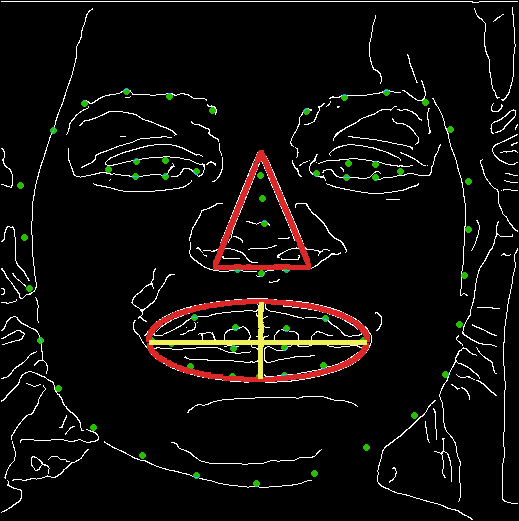}}\\
(c) & (d) \\
\end{tabular}
\caption{Illustration of the methodology and a subset of geometrical features.  (a) Region of interest with the localised facial landmarks. (b)  Three distance features illustrated with red lines:  R1 (distance between eyes), R2 (distance between edges of the nostrils), R3, (distance between the corners of the mouth) and three baselines; B1 (temples), B2 (cheekbones), B3 (jaw). (c) Triangular region corresponding to the nose and the angle $\alpha$. (d) Ellipsoidal region corresponding to the mouth. }
\label{fig:figure1b}
\end{figure}

4) {\bf Landmarks localization}: The localisation of landmarks follows the method of Kazemi \cite{Kazemi_Sullivan_2014}. In order to label the facial regions, a training data set  that consists of labelled facial landmarks on face images and  probability of distance between pairs of pixels near facial landmarks is used. Using the training data set as described above, in order to localise facial landmarks, an ensemble of regression trees is trained to estimate the facial landmark positions using only pixel intensities. For this work, a configuration 68 landmarks distributed around the eyes, nose, mouth and contour of the face was selected and is illustrated in Fig.~\ref{fig:figure1}(b). The landmarks overlaid on the original facial image are shown in Fig.~\ref{fig:figure1b}(a). \\



5) {\bf Geometrical feature extraction}: 
Patients with PTHS present special facial features  related to the geometry of the face. Therefore, geometric features, specifically of distance, area and angles,  as defined by the relationship among the landmarks previously located, were explored to support diagnosis of  PTHS.  

Three distance features, $R1, R2,  R3$ were defined by the distances between the eyes (landmarks 39 and 40,  in Fig.~\ref{fig:figure1b}(b)), width of the nose (landmarks 31 and 35) and width of the mouth (landmarks 48 and 54). These features were normalised by the respective baselines $B1, B2, B3$ (the width of the face in three  different locations) so that they are invariant to scale, rotation and translation. For the first feature $R1$, the baseline $B1$ is the distance between the temples, for $R2$ the baseline $B2$ is the distance between the cheekbones and for $R3$ the baseline $B3$ is the width of the jaw (Fig.~\ref{fig:figure1b}(b)).  


One angle feature called $NoseAngle$  was extracted to describe the width of the noses of individuals in terms of angular extension (noted as $\alpha$ in Fig.~\ref{fig:figure1b}(c)). Starting from a triangle built using three landmarks of nose as apices, the Carnot's theorem has been used in order to calculate the top angle of the triangle.

Two area features were derived using polygons or circles defined on facial landmarks. First, the  ratio of nose area over face area ($RNose$) and second, the ratio of mouth area over face area ($RMouth$)  as illustrated in Fig.~\ref{fig:figure1b}(d).
The area of the face was calculated as the area of the ellipse constructed on the basis of the two axes where the minor axis is the face width (landmark 1 to 15), and major axis is the height of the face (landmark 8, the chin, and  centroid between the landmarks 19 and 24 (Fig.~\ref{fig:figure1}(b)). \\

6) {\bf Statistical Analysis}: The six geometric features ($R1/B1$, $R2/B2$, $R3/B3$, $NoseAngle$, $Rnose$, $RMouth$) were compared between healthy and PTHS populations with a paired Student's t-test. \\

\begin{figure}[!ht]
\centering
\begin{tabular}{c}
 {\includegraphics[width=120mm,height=100mm]{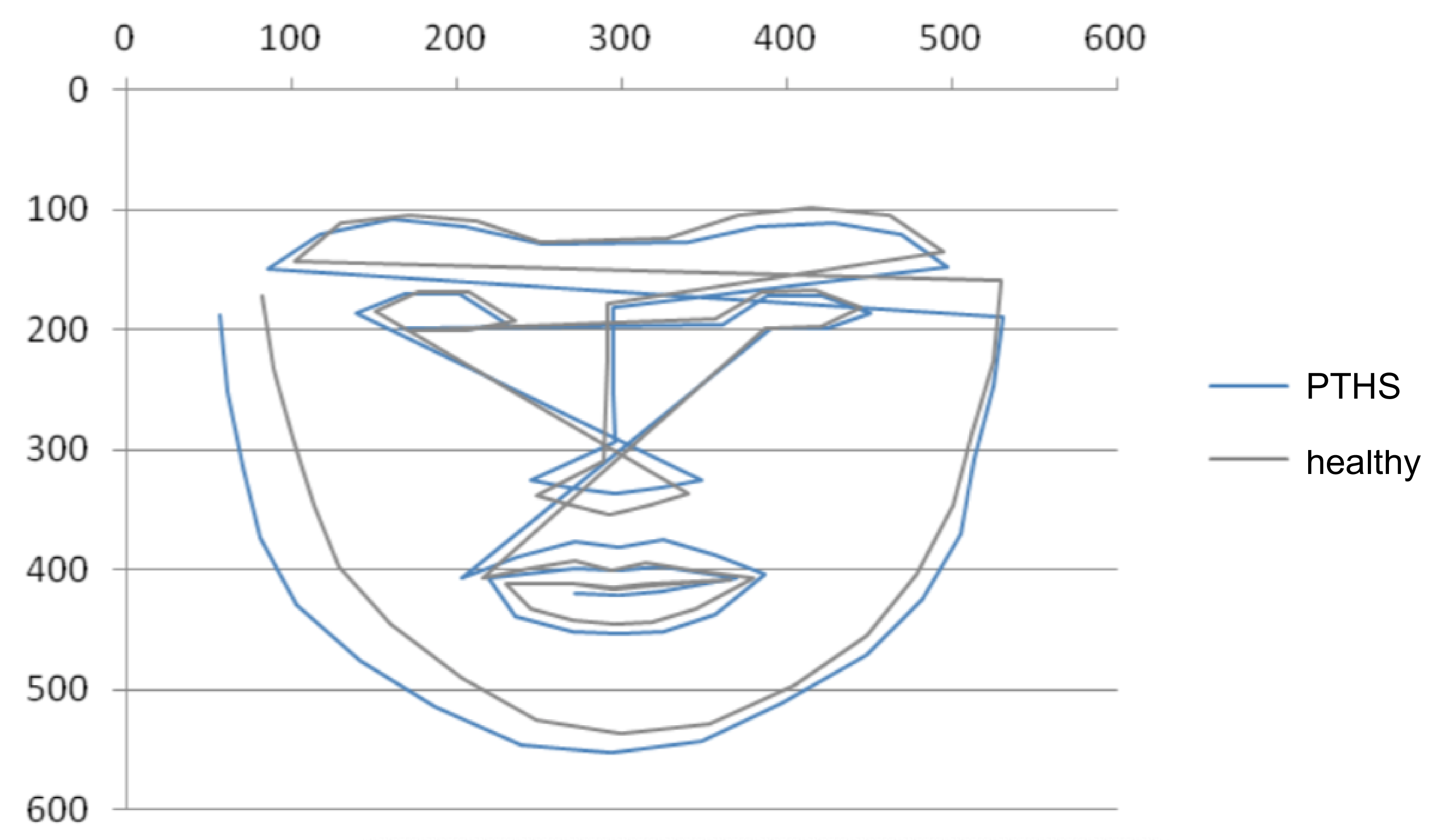}}\\
\end{tabular}
\caption{ Mean face shape comparison between PTHS and healthy individuals. It should be noticed the similarity in the regions corresponding to the eyes and the differences in the nose and mouth.  }
\label{fig:figure2}
\end{figure}

\begin{figure}[!ht]
\centering
\begin{tabular}{cc}
 (a) & {\includegraphics[width=105mm]{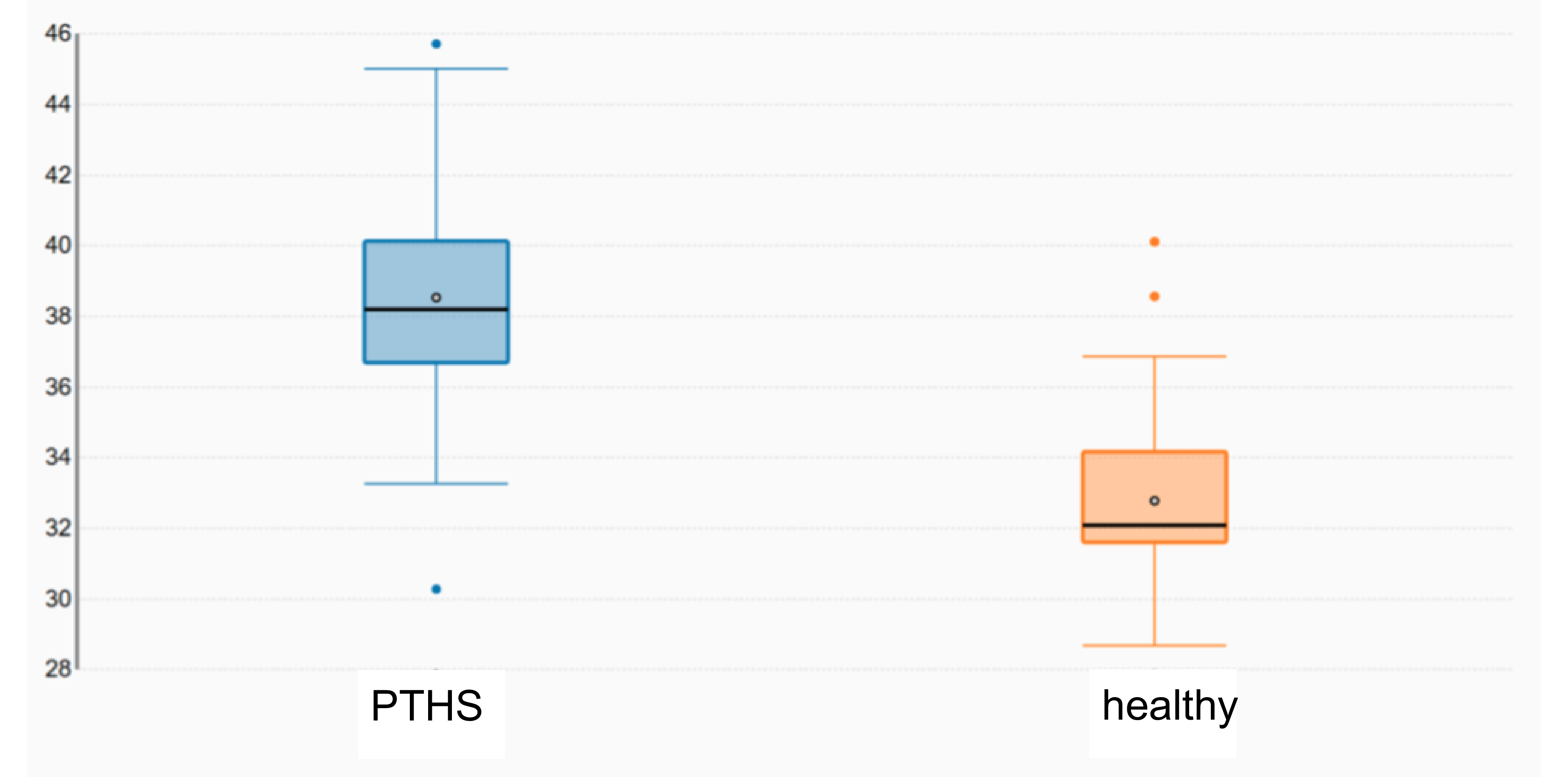}}\\
 (b) & {\includegraphics[width=120mm]{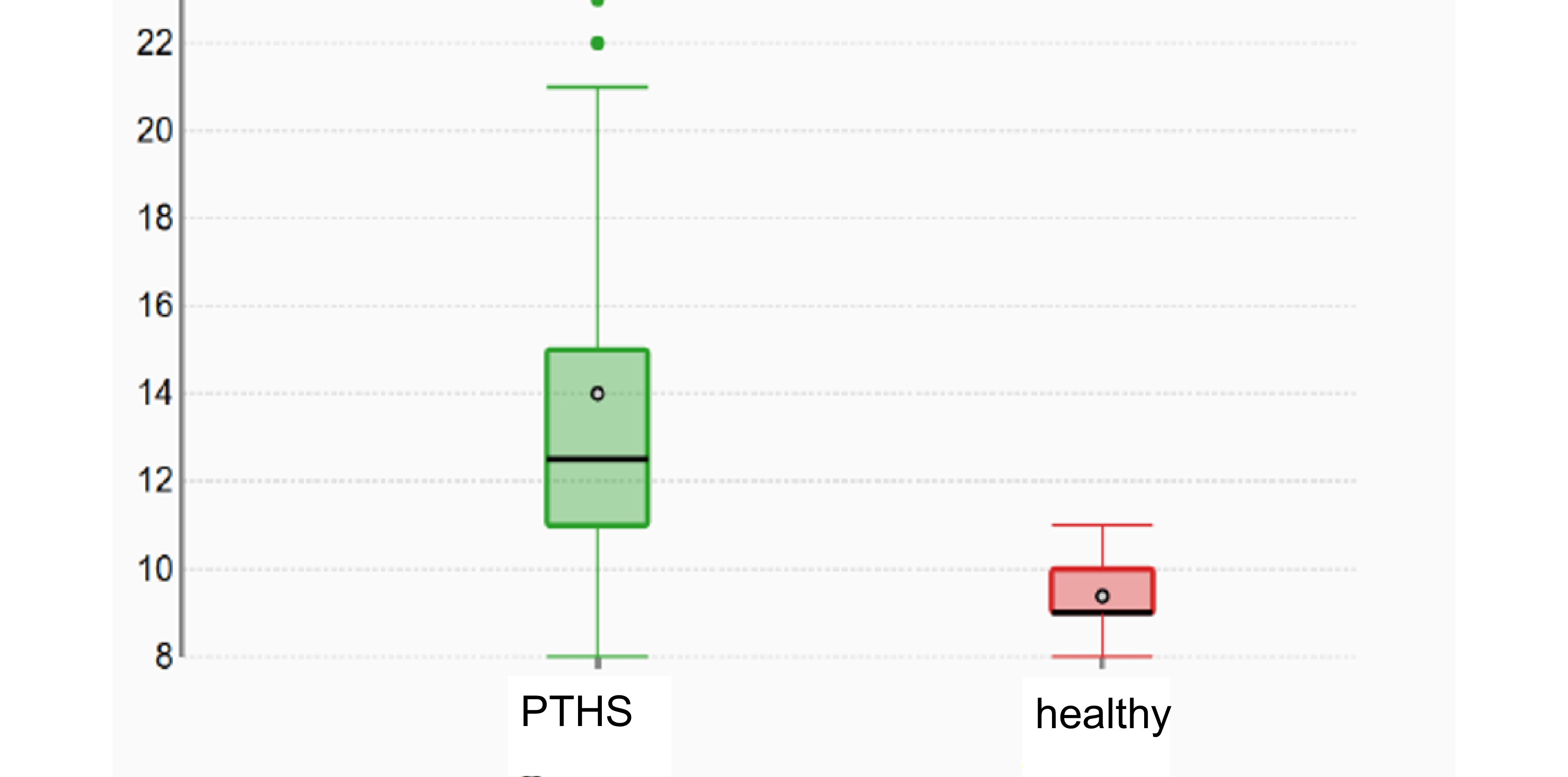}}\\
 
\end{tabular}
\caption{ Boxplots corresponding to the measurements of the feature  $NoseAngle$ and $RMouth$, both of which indicate a statistical difference ($p<$0.0001) between the populations. As expected, the nose and mouth of the PTHS group are wider than the healthy controls. }
\label{fig:figure3}
\end{figure}

\section{Results}
The methodology was applied to all images and the localised landmarks were assessed visually. None of the face images of the collected data set showed incorrectly localised landmarks. As general comparison between the two populations, the average location of the landmarks was calculated and displayed as a {\it mean face} (Fig.~\ref{fig:figure2}). Whilst the landmarks of the eyes and forehead seem to be close between populations, those of the nose, mouth and perimeter of the face were more distant to each other. In terms of statistical difference, it was found that there was no significant differences for the distances
$R1/B1$, $R2/B2$, $R3/B3$, nor for the area of the nose $Rnose$. However, there was a statistical differenve for the angle $NoseAngle$ and the ratio of the mouth $RMouth$. These differences are shown in Fig.~\ref{fig:figure3} as boxplots and  Table~\ref{table:table1} shows the $p$-values of the t-tests for all cases.


\begin{table}[]
    \centering
    \begin{tabular}{|c|c|}
    \hline
        Feature     & $p$-value \\
        \hline \hline
        $R1/B1$     &    0.446     \\
        \hline
        $R2/B2$     &  0.709     \\
        \hline
        $R3/B3$     &  0.185     \\
        \hline
        $NoseAngle$ &  $<$0.0001     \\
        \hline
        $Rnose$     &   0.68     \\
        \hline
        $RMouth$    &   $<$0.0001     \\
        \hline
    \end{tabular}\vspace{0.5cm}
    \caption{Statistical comparison between PTHS and healthy populations.
    }
    \label{table:table1}
\end{table}{}

\section{Conclusions}
The methodology described in this paper found a statistical difference between populations of healthy individuals and those with PTHS. The methodology is automatic and only requires frontal facial images without obstructions, but there is no user intervention in the process. Whilst the sample of the data is relatively small (PTHS 71, Healthy 55), the results are promising and could be used by health professionals interested in the recognition of the syndrome without the need of genetic tests. 

\vskip3pt
\vskip3pt


\bibliographystyle{splncs03_unsrt}
\bibliography{references}

\begin{thebibliography}{10}
\providecommand{\url}[1]{\texttt{#1}}
\providecommand{\urlprefix}{URL }

\bibitem{Marangi_Zollino_2015}
Marangi, G., Zollino, M.: Pitt-hopkins syndrome and differential diagnosis: A
  molecular and clinical challenge. Journal of Pediatric Genetics  4(3),
  168–176 (Sep 2015)

\bibitem{Goodspeed_Newsom_Morris_Powell_Evans_Golla_2018}
Goodspeed, K., Newsom, C., Morris, M.A., Powell, C., Evans, P., Golla, S.:
  Pitt-hopkins syndrome: A review of current literature, clinical approach, and
  23-patient case series. Journal of Child Neurology  33(3),  233–244 (2018)

\bibitem{Amiel_Rio_de}
Amiel, J., Rio, M., de~Pontual, L., Redon, R., Malan, V., Boddaert, N., Plouin,
  P., Carter, N.P., Lyonnet, S., Munnich, A., et~al.: Mutations in tcf4,
  encoding a class i basic helix-loop-helix transcription factor, are
  responsible for pitt-hopkins syndrome, a severe epileptic encephalopathy
  associated with autonomic dysfunction. American Journal of Human Genetics
  80(5),  988–993 (May 2007)

\bibitem{Brockschmidt_Todt_Ryu_Hoischen_Landwehr_Birnbaum_Frenck_Radlwimmer_Lichter_Engels}
Brockschmidt, A., Todt, U., Ryu, S., Hoischen, A., Landwehr, C., Birnbaum, S.,
  Frenck, W., Radlwimmer, B., Lichter, P., Engels, H., et~al.: Severe mental
  retardation with breathing abnormalities (pitt-hopkins syndrome) is caused by
  haploinsufficiency of the neuronal bhlh transcription factor tcf4. Human
  Molecular Genetics  16(12),  1488–1494 (Jun 2007)

\bibitem{Sweetser_Elsharkawi_Yonker_Steeves_Parkin_Thibert_1993}
Sweetser, D.A., Elsharkawi, I., Yonker, L., Steeves, M., Parkin, K., Thibert,
  R.: Pitt-Hopkins Syndrome. University of Washington, Seattle (1993),
  \url{http://www.ncbi.nlm.nih.gov/books/NBK100240/}

\bibitem{Kuru_Niranjan_Tunca_Osvank_Azim_2014}
Kuru, K., Niranjan, M., Tunca, Y., Osvank, E., Azim, T.: Biomedical visual data
  analysis to build an intelligent diagnostic decision support system in
  medical genetics. Artificial Intelligence in Medicine  62(2),  105–118 (Oct
  2014)

\bibitem{Samal_Iyengar_1992}
Samal, A., Iyengar, P.A.: Automatic recognition and analysis of human faces and
  facial expressions: a survey. Pattern Recognition  25(1),  65–77 (Jan 1992)

\bibitem{Schroff_Kalenichenko_Philbin_2015}
Schroff, F., Kalenichenko, D., Philbin, J.: Facenet: A unified embedding for
  face recognition and clustering. In: 2015 IEEE Conference on Computer Vision
  and Pattern Recognition (CVPR). vol.~1, p. 815–823. IEEE (2015)

\bibitem{Farnell_Galloway_Zhurov_Richmond_Perttiniemi_Katic_2017}
Farnell, D.J.J., Galloway, J., Zhurov, A., Richmond, S., Perttiniemi, P.,
  Katic, V.: Initial Results of Multilevel Principal Components Analysis of
  Facial Shape, p. 674–685. Communications in Computer and Information
  Science, Springer International Publishing (2017)

\bibitem{Gurovich_Hanani_Bar_Fleischer_Gelbman_Basel-Salmon_Krawitz_Kamphausen_Zenker_Bird}
Gurovich, Y., Hanani, Y., Bar, O., Fleischer, N., Gelbman, D., Basel-Salmon,
  L., Krawitz, P., Kamphausen, S.B., Zenker, M., Bird, L.M., et~al.:
  Deepgestalt - identifying rare genetic syndromes using deep learning.
  arXiv:1801.07637 [cs]  (Jan 2018), \url{http://arxiv.org/abs/1801.07637},
  arXiv: 1801.07637

\bibitem{King_2009}
King, D.E.: Dlib-ml: A machine learning toolkit. J. Mach. Learn. Res.  10,
  1755–1758 (Dec 2009),
  \url{http://dl.acm.org/citation.cfm?id=1577069.1755843}

\bibitem{Oliphant_2015}
Oliphant, T.E.: Guide to NumPy. CreateSpace Independent Publishing Platform,
  2nd edn. (2015)

\bibitem{vanderWalt_Schonberger}
van~der Walt, S., Schonberger, J.L., Nunez-Iglesias, J., Boulogne, F., Warner,
  J.D., Yager, N., Gouillart, E., Yu, T., scikit-image contributors:
  scikit-image: image processing in python. PeerJ  2,  e453 (2014)

\bibitem{Pedregosa_Varoquaux_Gramfort_Michel_Thirion_Grisel_Blondel_Prettenhofer_Weiss_Dubourg}
Pedregosa, F., Varoquaux, G., Gramfort, A., Michel, V., Thirion, B., Grisel,
  O., Blondel, M., Prettenhofer, P., Weiss, R., Dubourg, V.: Scikit-learn:
  Machine learning in python. J. Mach. Learn. Res.  12,  2825–2830 (Nov
  2011), \url{http://dl.acm.org/citation.cfm?id=1953048.2078195}

\bibitem{Viola_Jones_2004}
Viola, P., Jones, M.J.: Robust real-time face detection. International Journal
  of Computer Vision  57(2),  137–154 (May 2004),
  \url{https://doi.org/10.1023/B:VISI.0000013087.49260.fb}

\bibitem{Dalal_Triggs_2005}
Dalal, N., Triggs, B.: Histograms of oriented gradients for human detection.
  In: 2005 IEEE Computer Society Conference on Computer Vision and Pattern
  Recognition (CVPR’05). vol.~1, p. 886–893 vol. 1 (Jun 2005)

\bibitem{Deniz_Bueno_Salido_DelaTorre_2011}
D\'{e}niz, O., Bueno, G., Salido, J., De~la Torre, F.: Face recognition using
  histograms of oriented gradients. Pattern Recognition Letters  32(12),
  1598–1603 (Sep 2011),
  \url{http://www.sciencedirect.com/science/article/pii/S0167865511000122}

\bibitem{Cortes_Vapnik_1995}
Cortes, C., Vapnik, V.: Support-vector networks. Machine Learning  20(3),
  273–297 (Sep 1995), \url{https://doi.org/10.1007/BF00994018}

\bibitem{Kazemi_Sullivan_2014}
Kazemi, V., Sullivan, J.: One millisecond face alignment with an ensemble of
  regression trees. In: Proceedings of the 2014 IEEE Conference on Computer
  Vision and Pattern Recognition. p. 1867–1874. CVPR ’14, IEEE Computer
  Society (2014), \url{https://doi.org/10.1109/CVPR.2014.241}

\end{thebibliography}

\end{document}